\documentclass[journal]{IEEEtran}
\usepackage{soul} 
\usepackage{cite}
\usepackage{amsmath,amssymb,mathtools}
\usepackage{array}
\usepackage{stfloats}
\usepackage{algorithm,algpseudocode}
\usepackage[pdftex]{graphicx}
\usepackage{lipsum}
\usepackage{multirow}
\usepackage{textcomp}
\usepackage{url}
\graphicspath{{pics/}}
\DeclareGraphicsExtensions{.pdf,.jpeg,.png}
\ifCLASSOPTIONcompsoc
  \usepackage[caption=false,font=normalsize,labelfont=sf,textfont=sf]{subfig}
\else
  \usepackage[caption=false,font=footnotesize]{subfig}
\fi
\usepackage{color}
\usepackage{hyperref}
\algnewcommand\INPUT{\item[\textbf{Input:}]}%
\algnewcommand\OUTPUT{\item[\textbf{Output:}]}%

\DeclareMathOperator*{\argmin}{\arg\!\min}
\newcommand\norm[1]{\left\lVert#1\right\rVert}

\begin{document}
\title{A Tensor Factorization Method for 3D
 Super-Resolution\\ with Application to Dental CT
}

\author{Janka Hatvani, Adrian Basarab, Jean-Yves Tourneret, Mikl\'os Gy\"ongy and Denis Kouam\'e 
\thanks{This work has been submitted to the IEEE for possible publication. Copyright may be transferred without notice, after which this version may no longer be accessible.}
\thanks{This research has been partially supported by the European Union, co-financed by the European Social Fund (EFOP-3.6.3-VEKOP- 16-2017-00002 and by P\'azm\'any University KAP17-19.}
\thanks{This work was supported by the thematic trimester on image processing of the CIMI Labex, Toulouse, France, with the Program ANR-11-IDEX-0002-02 under Grant
ANR-11-LABX-0040-CIMI.
}
\thanks{J. Hatvani, A. Basarab, and D. Kouam\'e are with IRIT, CNRS UMR 5505, Université Paul Sabatier Toulouse 3, University of Toulouse, France.}
\thanks{J. Hatvani, and M. Gy\"ongy are with the Faculty of Information Technology and Bionics,
Pazmany Peter Catholic Unviersity, Budapest Hungary}
\thanks{J. Y. Tourneret is with  IRIT/INP-ENSEEIHT/T\'esa, University of Toulouse, France.}}

\markboth{}%
{Hatvani \MakeLowercase{\textit{et al.}}:
A Tensor Factorization Method for 3D Super-Resolution with application to Dental CT
}
\maketitle

\begin{abstract}

Available super-resolution techniques for 3D images are either computationally inefficient prior-knowledge-based iterative techniques or deep learning methods which require a large database of known low- and high-resolution image pairs. A recently introduced  tensor-factorization-based approach offers a fast solution without the use of known image pairs or strict prior assumptions. 
In this article this factorization framework is investigated for single image resolution enhancement  with an off-line estimate of the system point spread function. The technique is applied to 3D cone beam computed tomography for dental image resolution enhancement. To demonstrate the efficiency of our  method, it is compared to a  recent state-of-the-art  iterative technique using low-rank and total variation regularizations. 
In contrast to this comparative technique, the proposed reconstruction technique gives a 2-order-of-magnitude improvement in running time -- 2 minutes compared to 2 hours for a dental volume of 282$\times$266$\times$392 voxels. Furthermore, it also offers slightly improved quantitative results (peak signal-to-noise ratio, segmentation quality). Another advantage of the presented technique is the low number of hyperparameters. As demonstrated in this paper, the framework is not sensitive to small changes of its parameters, proposing an ease of use.

\end{abstract}


\section{Introduction}

\IEEEPARstart{R}{oot} canal treatment is carried out on a regular basis in dental centers in order to save decayed and infected teeth. In spite of their popularity, the success rate of the treatment is only 60-85\% \cite{ng2007outcome,eriksen2002endodontic}.
For an improvement of the process the dentists need a better visualization of the canal, as its length, diameter and curvature are all important factors for planning the therapy \cite{peters2004current}. Therefore, further research on the visualization possibilities of the pulp cavity is necessary, as stated by the European Commission on Radiation Protection in 2012 \cite[pp. 61-65]{sedentexct}.

Dental offices use cone beam computed tomography (CBCT) for determining the 3D structure of the teeth. Apart from the detector size, the spatial resolution of such imaging devices is also affected by partial volume effect, noise, and beam hardening, resulting in a typical value of 500 \textmu{}m. This resolution is not sufficient in endodonty since the diameter of the canal is usually in the range of 0.16-1.60 mm and the apical, narrower segment is more important for planning the treatment \cite{martos2011anatomical}.
On the other hand, the resolution of micro-CT (\textmu{}CT) is sufficient for precise measurements on the cavity, but the physical dimensions of the system only permit the imaging of extracted teeth. The long acquisition time and high radiation dose also prohibit \textit{in vivo} measurements. 

The post-processing application of image super-resolution (SR) algorithms is an intensely investigated field in the image processing community. Classical SR techniques can combine information from a sequence of measurements \cite{farsiu2004fast}, from different modalities \cite{yin2013simultaneous}, or in the simplest case they try to improve the resolution of a single image \cite{kim2010single}. Many SR methods assume that the low resolution (LR) image of interest is obtained from the high resolution (HR) image by blurring and decimation with a residual additive noise. The SR problem can then be formulated as an inverse problem, which is ill-posed and thus requires an appropriate regularization to provide suitable solutions.A standard regularization often employed is total variation (TV) leading to piecewise smooth solutions \cite{toma2014total}. Low-rank \cite{shi2015lrtv}, or wavelet representations \cite{zhang2018limited} have  also proved to be efficient tools for SR. A method based on a sparse representation was applied to 3D MRI images with a patch-based structural similarity constraint in \cite{manjon2010non}.  Convolutional neural networks have also shown interesting properties for SR, where the network is trained to map an LR image to its HR counterpart \cite{Oktay16,Cengiz17,Zhang17,hatvani2018deep}. However, this technique requires a large training dataset, which is not always available. 
Furthermore, only a few of the above-mentioned techniques is available for 3D volumes (\textit{e.g.,} \cite{shi2015lrtv,manjon2010non,Oktay16}), and they all suffer from heavy computational costs, preventing them from application in practice.

A new hyperspectral-multispectral image fusion technique using tensor factorization (TF) was introduced in \cite{Kanatsoulis18}. This technique combines a multispectral image (with high spatial and low spectral resolutions) and a hyperspectral image (with low spatial and high spectral resolution) to obtain an SR image (with high spatial and high spectral resolutions). 
One advantage of the tensor-based method of \cite{Kanatsoulis18} is that it does not need to unfold the image of interest into a 2D matrix as in many existing SR methods \cite{shi2015lrtv},\cite{hatvani2018deep}. As a consequence, this method avoids any loss of information about the locality of the image pixels and does not require to introduce spatial regularization (such as the TV of the image).

This paper investigates a 3D single image SR (SISR) method based on TF for CBCT images of teeth as an attempt to approximate their HR \textmu{}CT pairs. The idea is to decompose the image of interest using its canonical polyadic decomposition (CPD). The CPD of a tensor is a representation based on a sum of an appropriate number of rank-1 tensors, which number depends on the structure of the image. It will be shown in this paper that this representation leads to a notably fast and efficient reconstruction method. The described method is compared to a state-of-the-art iterative deconvolution technique with low-rank and TV regularization (LRTV). For validation the peak signal-to-noise ratio (PSNR) is calculated and the canal is segmented, permitting volumetric and diametric comparison.  

The rest of this paper is organized as follows. First, the tensor operations used in this paper are defined and a connection between image complexity and tensor decomposition is drawn. In Section \ref{section_methods} the proposed TF-SISR method is first defined for the CBCT resolution enhancement problem, followed by the data acquisition and the estimation of the blurring point spread function (PSF), ending with the introduction of the evaluation metrics. Section \ref{section_results} compares the images obtained by the two different SISR methods and discusses the possibilities and limits of tensor factorization. Finally a conclusion is drawn about the applicability of the introduced SR technique to dental imaging with some possible future work.

\section{Tensors and Image Complexity} \label{tensor}

\subsection{Notations}
For easier distinction, 2D matrices are denoted using uppercase letters (e.g., $A$) and 3D tensors by bold uppercase letters (e.g., $\boldsymbol{A}$). The uppercase letter with an overline (eg., $\overline{A}$) denotes a set of 2D matrices. 

\subsection{Factorization, mode product, matricization}

In this section, operations from tensor algebra necessary for the proposed method are summarized. Readers may refer to \cite{Kanatsoulis18} and \cite{kolda2009tensor} for further details.

A tensor is a generalization of vectors and matrices, where the order of the tensor indicates the dimensionality. A 3D CT image volume is a third-order tensor $\boldsymbol{X} \in \mathbb{R}^{I\times J\times K}$ from which one dimensional fibers can be extracted. Depending on which indices are fixed, there are mode-1 fibers denoted as $\boldsymbol{X}(:,j,k)$ vectors (columns), mode-2 fibers denoted as $\boldsymbol{X}(i,:,k)$ vectors (rows) and mode-3 fibers denoted as $\boldsymbol{X}(i,j,:)$ vectors. The outer product (denoted by $\circ$)  between one mode-1, one mode-2 and one mode-3 array forms a rank-1 third order tensor, written as
\begin{equation}
\label{outer}
\begin{split}
&\boldsymbol{X} = u\circ v\circ w, \\
&u\in\mathbb{R}^{I},v\in\mathbb{R}^{J} , w\in\mathbb{R}^{K}, \boldsymbol{X}\in\mathbb{R}^{I\times J\times K} \\
&\textnormal{where}\\
&\boldsymbol{X}(i,j,k) = u(i) v(j) w(k).
\end{split}
\end{equation}

The smallest number of rank-1 tensors that can sum up to form the tensor $\boldsymbol{X}$ is called the tensor rank of $\boldsymbol{X}$, denoted by $F$. The resulting factorization of $\boldsymbol{X}$ is called the CPD of $\boldsymbol{X}$ expressed as
\begin{equation}
\label{build}
\begin{split}
&\boldsymbol{X} = \sum_{f=1}^F U^1{(:,f)} \circ U^2{(:,f)} \circ U^3{(:,f)}\\
&\textnormal{where}\\
& \boldsymbol{X}(i,j,k) =  \sum_{f=1}^F U^1{(i,f)}  U^2{(j,f)}  U^3{(k,f)}.
\end{split}
\end{equation}

\begin{figure}[h]
\centering{
\includegraphics[width=3in]{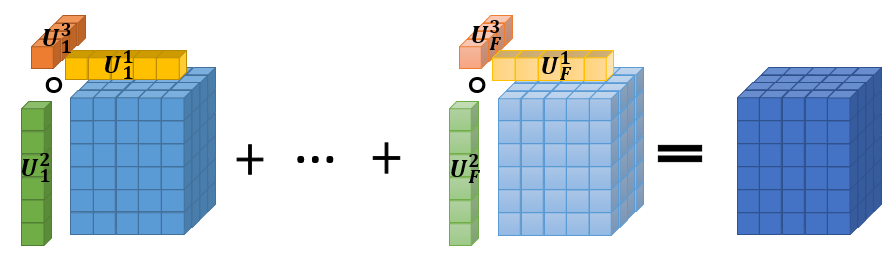}
\caption{Illustration of tensor factorization. $F$ is the number of outer products formed by mode-1 ($U^1_i := U^1(:,i)$), mode-2 ($U^2_i := U^2(:,i)$) and mode-3 ($U_i^3 := U^3(:,i)$) fibers summing up to a rank-$F$ tensor. 
}\label{factorize}}
\end{figure}

\noindent $\overline{U} = \left\lbrace {U^1},{U^2},{U^3} \right\rbrace$
is a set of three 2D matrices, $ \left\lbrace {U^1}\in\mathbb{R}^{I\times F}, {U^2}\in\mathbb{R}^{J\times F}, {U^3}\in\mathbb{R}^{K\times F} \right\rbrace$, known as the decomposition of the tensor $\boldsymbol{X}$.
For illustration, the reader may refer to Fig. \ref{factorize}. In the following, the operation in \eqref{build} will be denoted as
\begin{equation}
\label{build_notat}
 [\![{U^1}, {U^2}, {U^3}]\!]  = \sum_{f=1}^F U^1{(:,f)} \circ U^2{(:,f)} \circ U^3{(:,f)}.
\end{equation}
\noindent

An important property of the CPD is that this decomposition is essentially unique (allowing permutations within $U^1, U^2, U^3$). Thus $\overline{U}$ can be identified almost surely if its tensor rank $F$ is smaller than an upper bound. Chiantini \textit{et al.} \cite{chiantini2012generic} proved that if $ I \geq J \geq K$, with $F \leq 2^{\lfloor log_2J \rfloor +\lfloor log_2K \rfloor - 2}$, the CPD of the rank-$F$ tensor $\boldsymbol{X}\in \mathbb{R}^{I\times J\times K}$ is essentially unique. This condition allows identifiability of the CPD even for tensors with high rank. For example, in the application addressed herein, a typical CBCT volume with $260\times 260\times 300$ pixels can be decomposed uniquely even if the tensor rank of the image is as high as $2^{14}$ = 16384.

\begin{figure}[h]
\centering{
\includegraphics[width=3in]{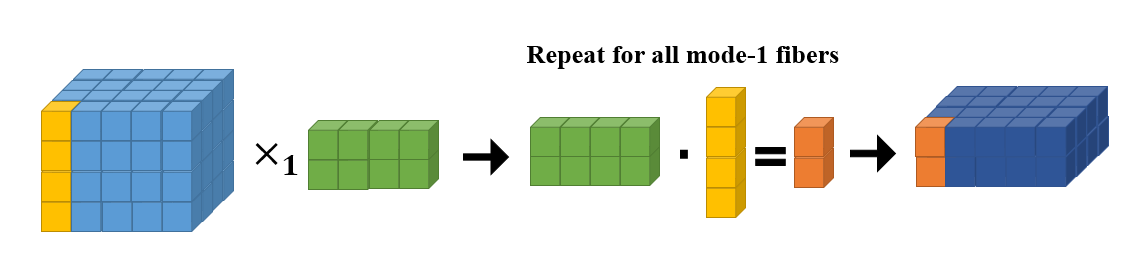}
\caption{Illustration of the mode-1 product. The mode-1 fibers of the 3D tensor are extracted and pre-multiplied by the 2D matrix. This example can illustrate a downsample operation with rate 2 in the first dimension.
}
\label{mode_prod}}
\end{figure}

Next, the multiplication between a 2D matrix and a 3D tensor referred to as the mode product is defined. This multiplication can be performed along all three dimensions, and in each case the mode-$n$ fibers of the tensor are extracted and are pre-multiplied by the matrix one-by-one. The mode-$n$ products ($n \in \left\lbrace1,2,3\right\rbrace$) of $\boldsymbol{X}\in\mathbb{R}^{I\times J\times K}$ with $P_1\in \mathbb{R}^{I^*\times I} , P_2\in \mathbb{R}^{J^*\times J}, P_3\in \mathbb{R}^{K^*\times K}$ are denoted as $\times_n$, and are defined as
\begin{equation}
\begin{split}
\boldsymbol{X}\times_1 P_1 = &\boldsymbol{X_1} \in\mathbb{R}^{I^*\times J\times K}\\ 
\textnormal{where }& \boldsymbol{X_1}(:,j,k) =  P_1\boldsymbol{X}(:,j,k) \\
\boldsymbol{X}\times_2 P_2 = &\boldsymbol{X_2} \in\mathbb{R}^{I\times J^*\times K}\\ 
\textnormal{where }&\boldsymbol{X_2}(i,:,k)= P_2\boldsymbol{X}(i,:,k) \\
\boldsymbol{X}\times_3 P_3 =  &  \boldsymbol{X_3}\in\mathbb{R}^{I\times J\times K^*}\\
\textnormal{where }& \boldsymbol{X_3}(i,j,:) =P_3\boldsymbol{X}(i,j,:)\\
\end{split}
\end{equation}

\noindent  where $I^*,J^*,K^*$ are arbitrary integer numbers. In Fig. \ref{mode_prod}, the principle of the mode-1 product, $\boldsymbol{X}\times_1P_1 = \boldsymbol{X_1}$ is illustrated, where the columns of the tensor are premultiplied by $P_1$, leading to a shrinkage along the first dimension.  

\noindent Using the factorization of $\boldsymbol{X}$ in (\ref{build}) and (\ref{build_notat}) the mode-$n$ products can also be written as
\begin{equation}
\label{modeprod}
\boldsymbol{X}\times_1 P_1\times_2 P_2\times_3 P_3 = [\![ P_1 {U^1},P_2{U^2},P_3{U^3} ]\!].
\end{equation}

Finally, the matricization or unfolding of the tensor $\boldsymbol{X} \in \mathbb{R}^{I\times J\times K}$
from 3D to 2D is defined. Note that this matricization can be realized again along the three dimensions. For a mode-\textit{n} matricization the mode-\textit{n} fibers are extracted and form the columns of $\boldsymbol{X}^{(n)}$ in lexicographical order expressed as
\begin{equation}
\label{matri}
\begin{split}
\boldsymbol{X}^{(1)}\textnormal{=}[&\boldsymbol{X}(:,1,1),\boldsymbol{X}(:,2,1),\textnormal{...}\\
&\boldsymbol{X}(:,J,1), \boldsymbol{X}(:,1,2)\textnormal{...},\boldsymbol{X}(:,J,K)]\\
\boldsymbol{X}^{(1)}&\in\mathbb{R}^{I\times JK}\\
\boldsymbol{X}^{(2)}\textnormal{=}[&\boldsymbol{X}(1,:,1),\boldsymbol{X}(2,:,1),\textnormal{...}\\
&\boldsymbol{X}(I,:,1), \boldsymbol{X}(1,:,2)\textnormal{...},\boldsymbol{X}(I,:,K)]\\
\boldsymbol{X}^{(2)}&\in\mathbb{R}^{J\times IK}\\
\boldsymbol{X}^{(3)}\textnormal{=}[&\boldsymbol{X}(1,1,:),\boldsymbol{X}(2,1,:),\textnormal{...}\\
&\boldsymbol{X}(I,1,:),\boldsymbol{X}(1,2,:)\textnormal{...},\boldsymbol{X}(I,J,:)]\\
\boldsymbol{X}^{(3)}&\in\mathbb{R}^{K\times IJ}.\\
\end{split}
\end{equation}

\noindent The same operation can be realized using the decomposition $\overline{U}$ of $\boldsymbol{X}$. For this the Khatri-Rao product -- denoted as $\odot$ -- is necessary. It operates on matrices having the same number of columns, and calculates their column-wise Kronecker-product ($A\odot B = C,$ where $ A\in\mathbb{R}^{I\times F}, B\in\mathbb{R}^{J\times F}, C\in\mathbb{R}^{IJ\times F}  $). Using this notation the matricization can be written as
\begin{equation}
\label{matri2}
\begin{split}
\boldsymbol{X}^{(1)} = U^1(U^3\odot U^2)^T\\
\boldsymbol{X}^{(2)} = U^2(U^3\odot U^1)^T\\
\boldsymbol{X}^{(3)} = U^3(U^2\odot U^1)^T.
\end{split}
\end{equation}

\subsection{Connection between the tensor rank and the image complexity} \label{connection}

\begin{figure}[h]
\centering{
\includegraphics[width=3.2in]{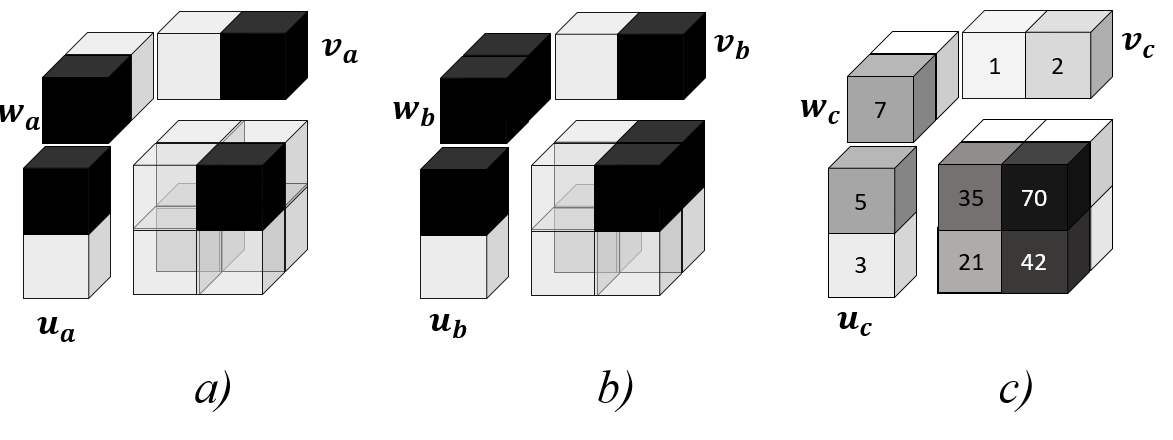}
\caption{Tensor rank and image complexity. In example \textit{a)} a single dark pixel (representing 1) in the white (representing 0) volume can be expressed by one outer product. In \textit{b)} two neighboring pixels are dark, making one outer product sufficient for their description. In \textit{c)} the pixel value is printed on the cell, equals 0 if not present. Two fibers are linearly dependent ($2\times$ [21,35] = [42,70]), so the volume can be decomposed using a tensor rank of $F=1$.
}\label{rank}}
\end{figure}

The tensor rank is related to the complexity of the image. To illustrate this claim, an illustrative set of examples is provided in Fig. \ref{rank}. In these examples the notations and dimensionality of  (\ref{outer}) are used with $I=J=K=2$. Fig. \ref{rank} \textit{a)} shows that the image with a single dark pixel (representing 1) in the white volume (representing 0) has a tensor rank of $F$ = 1. More precisely
\begin{equation}
\begin{split}
&u_a \circ v_a \circ w_a = [1,0] \circ [0,1] \circ[1,0] = \boldsymbol{X_a}\\
&\boldsymbol{X}^{(3)}_{\boldsymbol{a}} = 
\left[\begin{matrix}
0 &1& 0& 0  \\ 0& 0& 0& 0
\end{matrix}\right].
\end{split}
\end{equation}

In Fig. \ref{rank} \textit{b)}, two neighboring pixels are dark. This does not change the complexity of the image, since one outer product can still describe this volume. Indeed, we have
\begin{equation}
\begin{split}
&u_b \circ v_b \circ w_b = [1,0] \circ [0,1] \circ[1,1] =\boldsymbol{X_b}\\
&\boldsymbol{X}^{(3)}_{\boldsymbol{b}}  = 
\left[\begin{matrix}
0 &1& 0& 1  \\ 0& 0& 0& 0
\end{matrix}\right].
\end{split}
\end{equation}
In Fig. \ref{rank} \textit{c)}, two fibers are linearly dependent with $2\times$[21,35]=[42,70], which also makes one outer product sufficient for decomposing the tensor since
\begin{equation}
\begin{split}
&u_c \circ v_c \circ w_c = [5,3] \circ [1,2] \circ[7,0] = \boldsymbol{X_c}\\
&\boldsymbol{X}^{(3)}_{\boldsymbol{c}} = 
\left[\begin{matrix}
35 &70& 0& 0  \\ 21& 42& 0& 0
\end{matrix}\right].
\end{split}
\end{equation}

This set of illustrative examples shows that for images with piecewise constant regions (like the neighboring cells in Fig. \ref{rank} \textit{b)}) or with low matrix rank (as the linearly dependent fibers in Fig. \ref{rank} \textit{c)}) a smaller tensor rank can be expected. More generally, the tensor decomposition (\ref{build}) tends to promote solutions with small tensor ranks. This property is useful in the case of denoising, when independent outlier pixels have to be eliminated.  A degraded image may contain larger constant areas, with higher dependency between neighboring rows and columns. It means that describing these images will also be more efficient with a tensor of small rank. Thus these simple examples allow us to understand why CBCT images can be represented by a reduced number of rank-1 tensors.

\section{Methods} \label{section_methods}
\subsection{Problem formulation}

The image degradation model considered herein is the one classically used in SISR methods. It relates the LR image (CBCT in the case of the current dental application) to an HR image (considered to be close to the \textmu{}CT). The HR image $\boldsymbol{X}\in \mathbb{R}^{I\times J\times K}$ is corrupted by a decimation operator $D$ with rate $r$, a blurring kernel $H$, and some added noise $\boldsymbol{N}$,  resulting in the LR image $\boldsymbol{Y}\in \mathbb{R}^{I/r\times J/r\times K/r}$ such that
\begin{equation}
\begin{split}
\textnormal{vec}(\boldsymbol{Y}) = DH\textnormal{vec}(\boldsymbol{X}) + \textnormal{vec}(\boldsymbol{N})\\
\end{split}
\end{equation}

\noindent where vec($\cdot$) vectorizes the elements of the 3D tensor in lexicographical order. We assume that $H\in \mathbb{R}^{IJK\times IJK}$ is the block-circulant version of the 3D Gaussian kernel $h$  to avoid circular convolution. A 3D Gaussian kernel $h$ is separable along the three dimensions to 1D kernels as $h = h_1\circ h_2\circ h_3$ and is usually assumed for a blurring PSF \cite{elder1998local}. For Gaussian kernels $h_1,h_2,h_3$ with standard deviations $\sigma_1,\sigma_2,\sigma_3$ the corresponding block-circulant matrices are $H_1 \in \mathbb{R}^{I\times I},H_2\in \mathbb{R}^{J\times J},H_3\in \mathbb{R}^{K\times K}$.
The decimation operator downsamples the image by an integer number, by averaging blocks of $r$ neighboring pixels in each direction. In matrix form the downsampling operators for the three dimensions are $D_1 \in\mathbb{R}^{I/r\times I}, D_2\in\mathbb{R}^{J/r\times J}, D_3\in\mathbb{R}^{K/r\times K}$.
This formulation of $D$ corresponds for instance to the physical process of a large CBCT detector element collecting rays over a larger area, than \textmu{}CT does. This matrix also has better inversion properties compared to the regular decimation operator which discards pixels at a rate $r$.

Let $\overline{U} = \left\lbrace U^1 \in \mathbb{R}^{I\times F}, U^2 \in \mathbb{R}^{J\times F}, U^3\in\mathbb{R}^{K\times F} \right\rbrace$ be the CPD  of $\boldsymbol{X}$. 
The image degradation problem can be rewritten following (\ref{modeprod}) using the separated kernels
\begin{equation}
\begin{split}
\boldsymbol{Y} & = \boldsymbol{X}\times_1 D_1H_1\times_2 D_2H_2\times_3 D_3H_3 + \boldsymbol{N}\\
& = [\![ D_1H_1U^1,D_2H_2U^2,D_3H_3U^3 ]\!] + \boldsymbol{N}.
\end{split}
\end{equation}

\noindent The SR task can be defined as finding the set of matrices $\overline{U}$ which is the solution of the following minimization problem
\begin{equation}
\label{minim}
\min_{\overline{U}} \norm{\boldsymbol{Y} - [\![ D_1H_1U^1,D_2H_2U^2,D_3H_3U^3 ]\!]}_F^2
\end{equation}

\noindent where $\norm{\cdot}_F$ denotes the Frobenius norm of a tensor defined as the square root of the sum of its squared elements. This cost-function is different from the minimization problem of \cite{Kanatsoulis18} in the following aspects. First, only one measured datavolume is used here in contrast to the two measurements in the fusion problem of \cite{Kanatsoulis18}. Second, here the degradation happens in all three dimensions between the HR and LR image, while in \cite{Kanatsoulis18} the hyperspectral measurement is degraded in the first two dimensions, the multispectral volume in the third dimension. 

As problem \eqref{minim} is NP-hard an alternating optimization method is investigated, minimizing the cost function in \eqref{minim} sequentially for $U^1, U^2, U^3$.
Building a tensor from its decomposition  (\ref{build}) consists of a summation of $F$ outer products. Taking the square of this sum of $F$ terms in (\ref{minim}) would result in the sum of
${F+2-1}\choose{2}$ terms,
where ${\cdot}\choose{\cdot}$ denotes the binomial coefficient.
Instead when minimizing over $U_n$, the tensors are mode-\textit{n}-matricized using (\ref{matri}) and (\ref{matri2}) leading to
\begin{equation}
\begin{split}
\min_{U^1}\frac{1}{2} \norm{\boldsymbol{Y}^{(1)} - D_1H_1U^1(D_3H_3U^3\odot D_2H_2U^2)^T}_F^2\\
\min_{U^2}\frac{1}{2} \norm{\boldsymbol{Y}^{(2)} - D_2H_2U^2(D_3H_3U^3\odot D_1H_1U^1)^T}_F^2\\
\min_{U^3}\frac{1}{2} \norm{\boldsymbol{Y}^{(3)} - D_3H_3U^3(D_2H_2U^2\odot D_1H_1U^1)^T}_F^2.\\
\end{split}
\end{equation}

\noindent Note that the unfolding is performed in each direction sequentially, conserving the 3D local information. These minimizations can now be carried out easily, leading to
\begin{equation}
\label{solution}
\begin{split}
U^1 = (D_1H_1)^{+}\boldsymbol{Y}^{(1)}(D_3H_3U^3\odot D_2H_2U^2)^{+T}\\
U^2 = (D_2H_2)^{+}\boldsymbol{Y}^{(2)}(D_3H_3U^3\odot D_1H_1U^1)^{+T}\\
U^3 = (D_3H_3)^{+}\boldsymbol{Y}^{(3)}(D_2H_2U^2\odot D_1H_1U^1)^{+T}\\
\end{split}
\end{equation}

\noindent where $^+$ is the regularized Moore-Penrose pseudo-inverse defined as 
\begin{equation}
A^{+} = A^TA + \epsilon^2I)^{-1}A^T
\end{equation}

\noindent where $\epsilon$ is a hyperparameter used to provide a stable inverse (this procedure is classically referred to as diagonal loading).

To implement the proposed TF-SISR method the TensorLab toolbox \cite{tensorlab3.0} was used in Matlab 2017b. In the algorithm $\overline{U}$
was initialized with elements from the standard normal distribution and $U^1,U^2,U^3$ were updated iteratively several times as described in Algo. \ref{algo}.

\begin{algorithm}
    \caption{TF-SISR algorithm}
    \label{algo}
  \begin{algorithmic}[1]
    \INPUT $\boldsymbol{Y}\in\mathbb{R}^{I/r\times J/r\times K/r},F,[\sigma_1,\sigma_2,\sigma_3],r$
    \State \textbf{Initialize} $\overline{U} = \left\lbrace U^1 \in \mathbb{R}^{I\times F}, U^2 \in \mathbb{R}^{J\times F}, U^3\in\mathbb{R}^{K\times F} \right\rbrace$  with normally distributed values
    \State $ \begin{array}{l}
    D_1,D_2,D_3 \leftarrow \textnormal{decimation operator with a factor } r
    \end{array} $
    \State$\begin{array}{ll}H_1,H_2,H_3 \leftarrow &
    \textnormal{Gaussian kernels with standard} \\ 
    & \textnormal{deviations }[\sigma_1,\sigma_2,\sigma_3]   
    \end{array}$ 
    \While{stopping criteria is not met}   

      \State $U^1 \leftarrow Y^{(1)}, U^2, U^3$ 
      \State $U^2 \leftarrow Y^{(2)}, U^1, U^3$ \Comment{update using (\ref{solution})}
       \State $U^3 \leftarrow Y^{(3)}, U^1, U^2$      
    \EndWhile
    \State $\boldsymbol{X} \leftarrow \overline{U}$  \Comment{build using (\ref{build})}
    \OUTPUT $\boldsymbol{X}$, the estimated high resolution image
  \end{algorithmic}
\end{algorithm}

\subsection{Data acquisition}

The dataset used for testing contains images of 13 teeth which were extracted for health reasons and donated anonymously for research. This set consists of all different tooth types, including incisors, canines, premolars and molars.
A Carestream 81003D
system was used for CBCT imaging.
The linewidth resolution of the CBCT machine was 500 \textmu{}m and the volumes had a voxel size of $80\times 80\times 80$ \textmu{}m$^3$. 

For evaluation purposes the reconstructed HR images were compared to \textmu{}CT images acquired from the same samples.
The \textmu{}CT acquisitions were obtained with a Quantum FX system from Perkin Elmer, with a voxel size of $40\times 40\times 40$ \textmu{}m$^3$ and linewidth resolution of 50 \textmu{}m.



\subsection{PSF estimation}

The proposed method in Algo. \ref{algo} assumes that the PSF is known. In practice the blurring kernel has to be measured or estimated, which is usually carried out empirically in many existing works \cite{toma2014total}. Here the blurring kernel was assumed to be Gaussian and its standard deviation was estimated from the observed data. Employing direct inverse filtering on each sample image, the \textmu{}CT volume was divided by the CBCT volume in the frequency domain to obtain the Fourier transform of the PSF. The high-frequency band was suppressed by a Hanning-window before computing the inverse Fourier-transform and averaging the 13 estimated PSFs. For further details see \cite{hatvani2018deep}. Finally a 3D Gaussian function was fitted to the averaged PSF to estimate the standard deviations $\sigma_1, \sigma_2$ and $\sigma_3$.





\subsection{Metrics} \label{metri}


The comparison of the image volumes was carried out through two metrics.
The first one measures the PSNR between the estimated SR CBCT and \textmu{}CT volumes. It is calculated by dividing the square of the dynamic range with the mean square error between the enhanced image and the \textmu{}CT image, expressed in dB.

The second, more application oriented metric  consists of comparing volumes segmented from \textmu{}CT and SR CBCT. The canal root was segmented with a dedicated adaptive local thresholding method (see \cite{michetti2017comparison} for details). The canal area and Feret's diameter (the longest distance between two parallel straight lines that are tangent to the shape) were calculated for each radicular axial slice. The measured values are compared through the mean of absolute differences. The differences of the  canal volumes were also measured  using the Dice coefficient \cite{dice1945measures}.
Finally MeVisLab \cite{mevislab} was used for visualizing the segmentation results.



\section{Results} \label{section_results}

\subsection{Comparison to an existing 3D SR method}

The state-of-the-art LRTV introduced in 2013 in \cite{shi2013low,shi2015lrtv} was used as a benchmark to compare the performance of the proposed method\footnote{The Matlab code associated with LRTV is available at \href{https://bitbucket.org/fengshi421/superresolutiontoolkit}{https://bitbucket.org/fengshi421/superresolutiontoolkit}}.
 Among the relatively small collection of 3D SISR techniques, LRTV provided competitive results compared to other popular methods (cubic interpolation, non-local means, TV-based up-sampling) \cite{shi2015lrtv}.  It uses low-rank and total-variation regularizers, minimizing the following \\cost-function
\begin{equation}
\label{lrtv} 
\begin{split}
\boldsymbol{\hat{X}} = &\argmin_{\boldsymbol{X}} \norm{DH\boldsymbol{X}-\boldsymbol{Y}}^2 \\ 
&+ \lambda_{R}\textnormal{Rank}(\boldsymbol{X}) + \lambda_{TV}\textnormal{TV}(\boldsymbol{X}), 
\end{split}
\end{equation}
\noindent where $\lambda_{R}$ and $\lambda_{TV}$ are hyperparameters.
The minimization problem \ref{lrtv} is solved by the alternating direction method of multipliers (ADMM), which requires to adjust two additional hyperparameters, the penalty term $\rho$ and an iteration number $n_{\textnormal{ADMM}}$. One of the subproblems within the ADMM scheme is solved by gradient descent with an additional iteration number $n_{\textnormal{grad}}$ and an update rate denoted as $dt$.

The parameters used for testing can be seen in Table \ref{param}. They were tuned manually to get the highest possible improvement of the PSNR. The tests were run on a standard PC with an Intel(R) Core(TM) i7 2$\times$2.5GHz processor and 16 GB of RAM.

\begin{table}[h]
\caption{Parameters}
\label{param}
\centering
\begin{tabular}{|c|ccc|}
\hline
\multicolumn{2}{|c|}{LRTV}&\multicolumn{2}{|c|}{TF-SISR}\\
\hline\hline
\multicolumn{2}{|c|}{$n_{\textnormal{ADMM}}$ = 5}&\multicolumn{2}{|c|}{$n_{\textnormal{TF}}$ = 10}\\
\multicolumn{2}{|c|}{$\sigma = [5.8, 5.3, 0.9] $}&\multicolumn{2}{|c|}{$\sigma = [5.8, 5.3, 0.9] $}\\
\multicolumn{2}{|c|}{$\lambda_{TV} = 0.02 $}&\multicolumn{2}{|c|}{$F = 500$}\\
\multicolumn{2}{|c|}{$\lambda_{R} = 0.05$}&\multicolumn{2}{|c|}{$\epsilon = 1$}\\
\multicolumn{2}{|c|}{$\rho = 0.05$}&\multicolumn{2}{|c|}{}\\
\multicolumn{2}{|c|}{$n_{\textnormal{grad}}=100$}&\multicolumn{2}{|c|}{}\\
\multicolumn{2}{|c|}{$dt = 0.05$}&\multicolumn{2}{|c|}{}\\
\hline
\end{tabular}
\end{table}

The two methods were tested for three samples from the dataset, including an incisor, a premolar and a molar. The sizes of the sample volumes, the PSNR calculated against the \textmu{}CT images and the execution times  are provided in Table \ref{rtime}. Compared to the CBCT images the PSNR improves similarly for the LRTV (average of 1.2 dB) and the TF-SISR (average of 1.5 dB) methods  with the chosen parameters. However, this enhancement is achieved at a much lower computational cost: 10 iterations of TF-SISR run 100 times faster than 5 iterations of LRTV. This faster execution time is important since it permits a wider range of applications, including those requiring a rapid diagnosis during a medical examination.
In Fig. \ref{fig_bg}, the quality of the enhanced volumes is visualized, showing that the canal is better defined and contrasted compared to the CBCT image, suggesting better segmentation properties.

\textbf{\begin{figure}[t]
\fnbelowfloat
\centering
\includegraphics[width=3.5in]{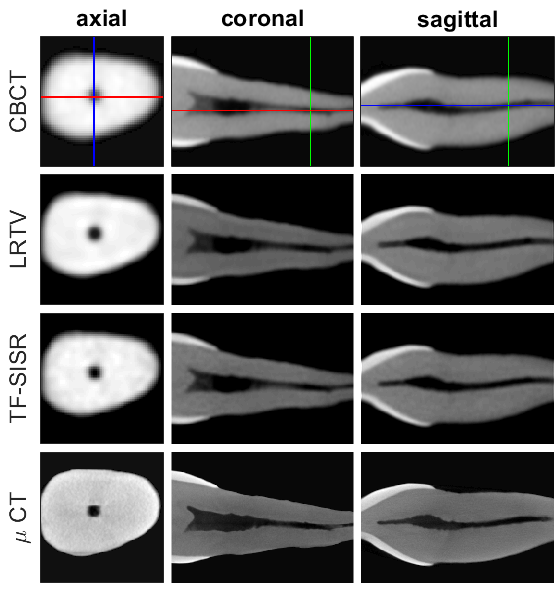}
\caption[]{Results on Sample \#1. In the rows the CBCT, LRTV output, TF-SISR output and \textmu{}CT images can be seen, whereas the columns correspond to one axial, a coronal and a sagittal slice. The CBCT image  is shown on the same scale as the HR images, for better comparison. The location of the slices within the volume is illustrated on the CBCT images in colored lines.}
\label{fig_bg}
\end{figure}}

\begin{table}[h]
\caption{Test results}
\label{rtime}
\centering
\begin{tabular}{|l|ccc|}
\hline
&Sample \#1&Sample \#2&Sample \#3\\
\hline\hline
tooth type&upper incisor&lower premolar& lower molar\\
\textmu{}CT image size&282$\times$266$\times$392&280$\times$268$\times$492&324$\times$306$\times$402\\
\hline
CBCT PSNR&23.17 dB&22.67 dB&24.14 dB \\
LRTV PSNR& 24.32 dB &24.65 dB& 24.61 dB\\
TF-SISR PSNR& 24.32 dB &24.48 dB& 25.71 dB\\
\hline
LRTV time&6988 s&9059 s& 10301 s\\
TF-SISR time& \textbf{71 s} &\textbf{92 s}&\textbf{104 s}\\
\hline
\end{tabular}
\end{table}

\begin{figure}[t]
\fnbelowfloat
\centering
\includegraphics[width=3.5in]{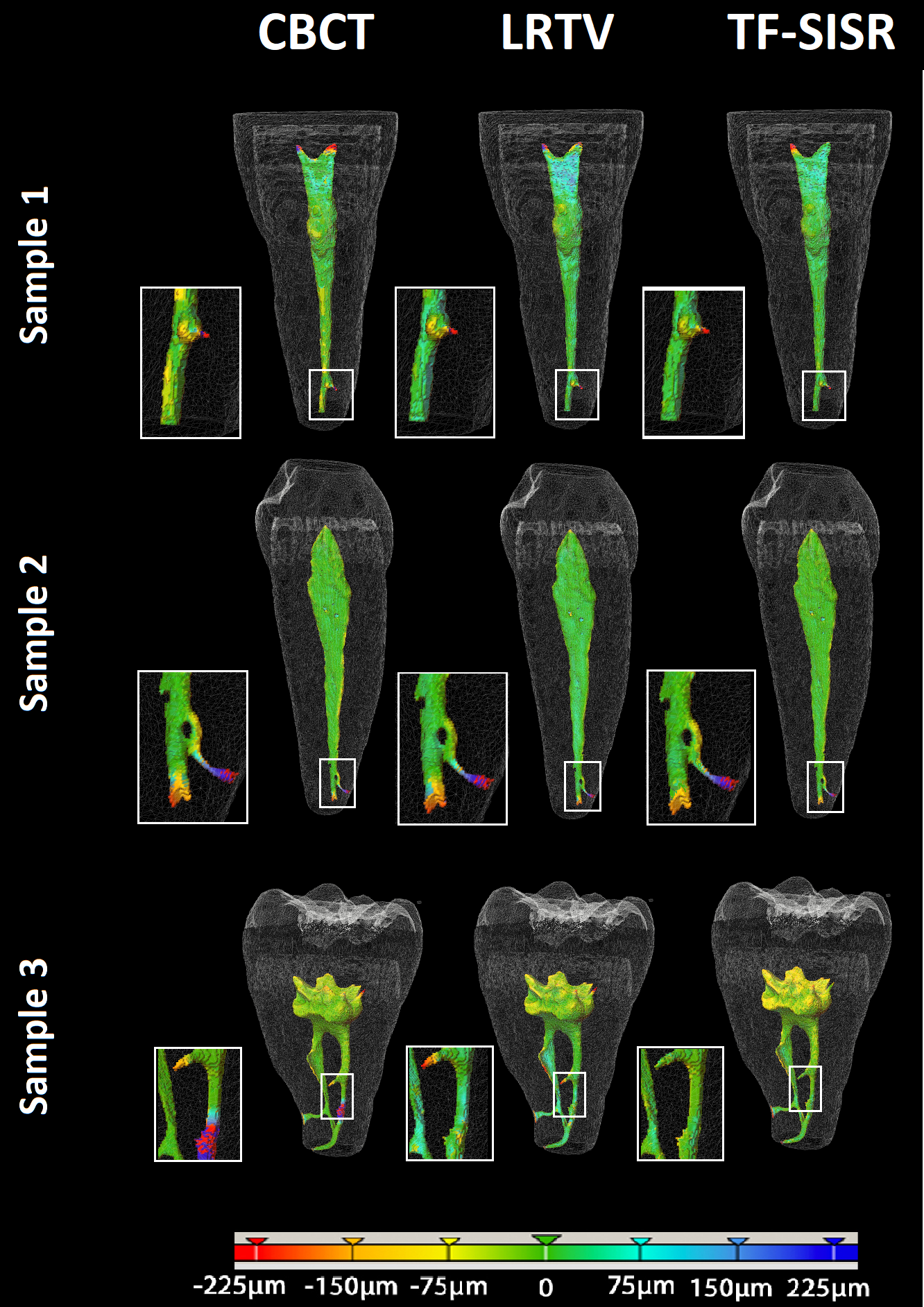}
\caption[]{Segmentation results for CBCT, LRTV and TF-SISR for the 3 samples. The color-bar visualizes the distance between the estimated surface of the canal and the one obtained with \textmu{}CT segmentation.}
\label{fig_seg}
\end{figure}

For further analysis the root canal was segmented from each volume, using the segmentation method described in Section \ref{metri}. Qualitative and quantitative results are provided in Fig. \ref{fig_seg} and in Table \ref{table_segmented}. In particular, Table \ref{table_segmented} shows differences between the estimates and the values obtained using the \textmu{}CT image for three parameters (Feret diameter, area of the canal and Dice coefficient). The estimated Feret diameter improves similarly with both SR techniques compared to the CBCT images with an averaged improvement of 63 \textmu m for LRTV and 81 \textmu m for TF-SISR. The second line of the table shows how the area of the canal on the axial slices is changing from one method to another. Note that the LRTV method shows a higher difference compared to the original CBCT (by 0.0256 mm$^2$), suggesting that the TV regularization overestimates the canal. This observation is also confirmed in Fig. \ref{fig_seg}, as the LRTV volumes have a more blueish color corresponding to positive differences. The TF-SISR provides the best overall performance with an improvement of 0.0152 mm$^2$ on average. The last metric in the table is the Dice-coefficient, also showing some improvement in the overlap of the canals, by 1\% using the LRTV and 2\% with the TF-SISR. Fig. \ref{fig_seg} displays zoomed-in sections of the apical part of the canal, as this part is the most important during the treatment. Considering these results, the TF-SISR method shows slightly better segmentation properties than the LRTV technique,  while offering a great reduction in running time.

\setlength{\tabcolsep}{4pt}
\begin{table}[t]
\centering
\renewcommand{\arraystretch}{1}
\caption{Canal Segmentation Metrics}
\label{table_segmented}
\begin{tabular}{|l|l|ccc|c|}
\hline
&method&Sample \#1&Sample \#2&Sample \#3& mean\\
\hline\hline
 {\multirow{3}{*}{\parbox{1.4cm}{\centering Mean of Diff. - Feret (\textmu{}m)}}} &
 CBCT &96&	89&	341&	176\\
 & LRTV &74&	71&	196&	113\\
 & TF-SISR &50&	57&	177&	95\\
 \hline
 {\multirow{3}{*}{\parbox{1.4cm}{\centering Mean of Diff. - Area (mm$^2$)}}}&
 CBCT &0.0463&0.0461&0.2492&0.1139\\
 & LRTV &0.0914& 0.0920& 0.2350&	0.1395\\
 & TF-SISR &0.0447&	0.0271&	0.2243&	0.0987\\
 \hline
 {\multirow{3}{*}{\parbox{1.4cm}{\centering Dice coefficient}}} &
 CBCT &0.88&	0.88&	0.90&	0.88\\
 & LRTV &0.87&	0.88&	0.90&	0.89\\
 & TF-SISR &0.90&	0.91&	0.91&	0.90\\
 \hline
\end{tabular}
\end{table}

\subsection{Adjusting the parameters of the TF-SISR method}

\begin{figure}[b]
\fnbelowfloat
\centering
\includegraphics[width=3.5in]{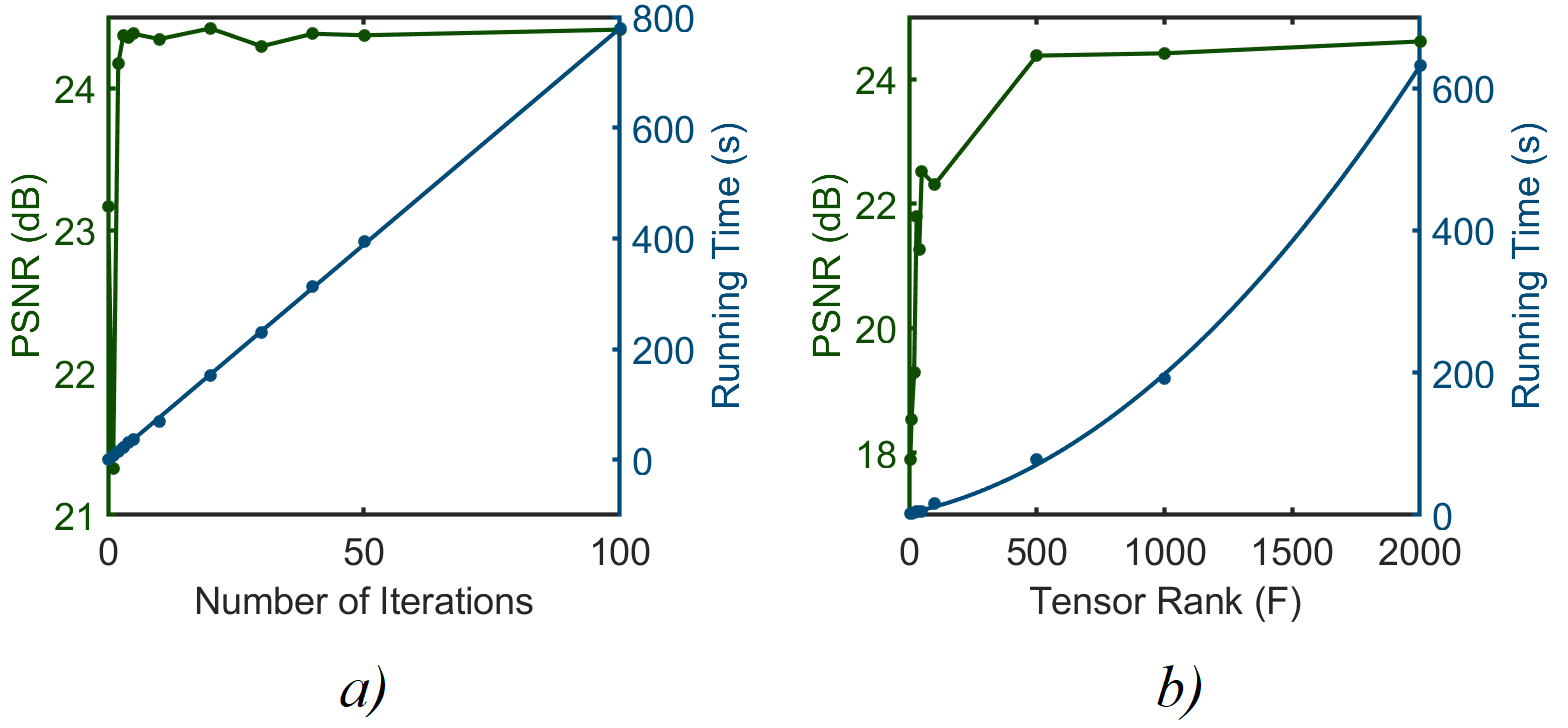}
\caption[]{Effect of the iteration number and the tensor rank on the PSNR and runtime. The rest of the parameters are as in Table \ref{param}. In \textit{a)} the PSNR saturates after a small number of iterations, while the runtime increases linearly. In \textit{b)} the runtime has am exponential growth versus the tensor rank and the PSNR saturates around $F = 500$.}
\label{evol_graph}
\end{figure}

\begin{figure*}[t]
\fnbelowfloat
\centering
\includegraphics[width=7.2in]{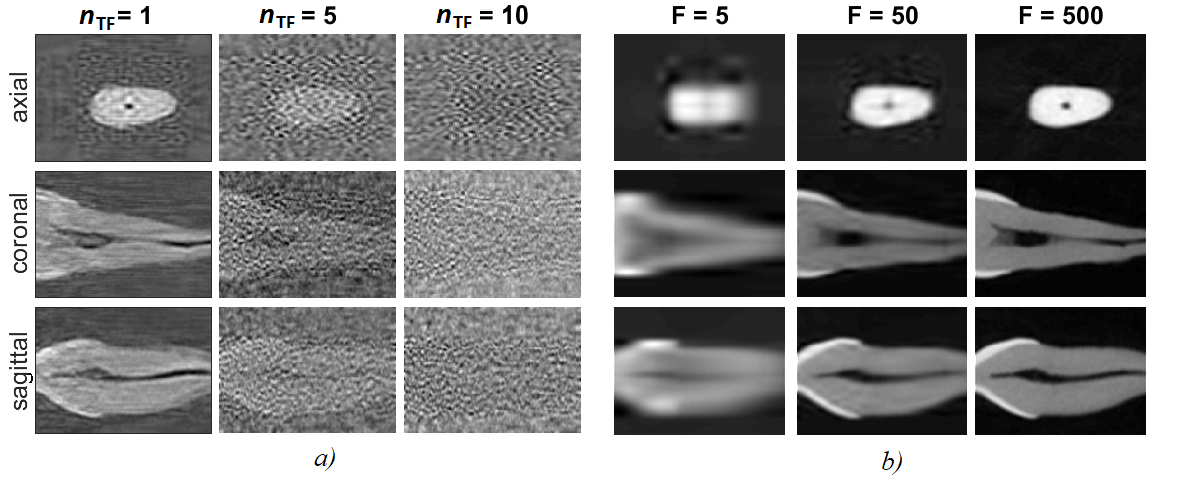}
\caption[]{Effect of the iteration number $it$ and the tensor rank $F$ in the reconstructed images. Sample \#1 is visualized through 3 slices from the axial, coronal and sagittal directions. In panel \textit{a)} the difference  compared to a result obtained after 10 iterations is shown. In case of $n_{\textnormal{TF}}$=10, a second test run was used for calculating the difference (note that the algorithm has random initialization, therefore different runs result in slightly different outputs). With more iterations the difference becomes less structured, more random. In panel \textit{b)} the change with $F$ can be seen: low numbers cause large blocks in the images, and the higher the tensor rank, the more detailed the output is. }
\label{evol_fig}
\end{figure*}

The impact of the tensor rank and the iteration number was investigated using Sample \#1.
Fig. \ref{evol_graph} \textit{a)} shows that the runtime increases linearly with the number of iterations, as expected.
Fig. \ref{evol_graph} \textit{a)} also shows that the PSNR converges rapidly to its maximum value (close to 24.5), which is an interesting property of the proposed method.
Fig. \ref{evol_fig} \textit{a)} shows how the solutions qualitatively evolve with the iteration number. For improved visibility the difference from the $n_{\textnormal{TF}} = 10$ case is shown in the figure. In the case of $n_{\textnormal{TF}} = 10$ a second test was run, and the difference was calculated compared to this result, as the random initialization of $\overline{U}$ results in slightly different outputs. It can be seen that as the iteration number increases, the difference becomes less structured: in the third column the shape of the tooth is almost invisible and it is lost in random noise.

According to the upper limit of $F$ for a unique CPD, in the case of Sample \#1, $F\le 2^{14}$ should be efficient. However, numbers higher than 2000 caused memory problems, and were therefore not tested. Fig. \ref{evol_graph} \textit{b)} shows that the computational time increases exponentially with the rank $F$ since the algorithm requires the inversion of larger matrices in $\overline{U}$. It can also be seen that the PSNR stabilizes for ranks larger than $F=500$. Some sample images can be seen in Fig. \ref{evol_fig} \textit{b)} showing that low values of the rank $F$ lead to large constant blocks in the image, which is characteristic of a low-rank or TV regularization. For higher numbers finer details become visible.

Note that our results indicate that neither of the above parameters have to be estimated precisely. After a small number of iterations the result converges. $F$ can be considered as a prior information on the complexity of the image. Using higher values a more natural result can be obtained, but above a threshold the method will not give more precise outputs.

\section{Conclusion}

In this work, a tensor-factorization-based method was proposed for the 3D single image super-resolution problem. This method showed interesting computational advantages compared to currently available regularization-based methods, with slightly improved image quality compared to the investigated LRTV technique. The runtime of this method was about 100 times faster than with LRTV, allowing a wider field of applications. The method also uses significantly less parameters (tensor rank and iteration number) that can be easily adjusted by visual inspection of the reconstruction results.
Dental CBCT volumes used as experimental data showed improved PSNR and canal-segmentation properties, with moderately better results than the LRTV method.  Considering these results, the method was found to be promising for 3D single image super-resolution.

The prior information in regularization-based techniques is often empirical and guides the solution. Future work can investigate if such classical priors could be included in this framework, and whether they would improve the result. As \cite{Kanatsoulis18} proposed a solution also for embedded PSF estimation, its application to TF-SISR  could be a potential direction of further research.

\section*{Acknowledgment}

The authors would like to thank J\'er\^ome Michetti for the data acquisition and the work on the visualization of the teeth.

\ifCLASSOPTIONcaptionsoff
  \newpage
\fi
\bibliographystyle{bib/IEEEtran}
\bibliography{bib/IEEEabrv,bib/dental}

\end{document}